\documentclass[default,iicol]{sn-jnl}

\usepackage{graphicx}
\usepackage{amsmath,amssymb}
\usepackage{multirow}
\usepackage{url}
\usepackage{xcolor}
\usepackage{bm}
\usepackage{mathrsfs}
\usepackage{pifont}
\usepackage{ragged2e}
\usepackage{makecell}
\usepackage{colortbl}
\usepackage{rotating}
\usepackage{dcolumn}
\usepackage{threeparttable}

\graphicspath{{./Imgs/},{./Imgs/Authors/}}
\DeclareGraphicsExtensions{.pdf,.jpg,.png}

\newcommand{\figref}[1]{Fig.~\ref{#1}}
\newcommand{\tabref}[1]{Table~\ref{#1}}
\newcommand{\secref}[1]{Sec.~\ref{#1}}

\newcommand{\equref}[1]{Eq.~(\ref{#1})}

\def\ie{\emph{i.e.}}

\def\vs{\emph{vs.~}}

\def\etal{{\em et al.~}}
\def\sArt{{state-of-the-art~}}

\definecolor{mygray}{gray}{.92}
\newcommand{\tabincell}[2]{\begin{tabular}{@{}#1@{}}#2\end{tabular}}
\newcommand{\myPara}[1]{\vspace{.12in}\noindent\textbf{#1}\quad}

\jyear{2021}%

\begin{document}

\title[HAT]{Vision Transformers with Hierarchical Attention}


\author[1]{\fnm{Yun} \sur{Liu}}

\author[2]{\fnm{Yu-Huan} \sur{Wu}}

\author[3]{\fnm{Guolei} \sur{Sun}}
\equalcont{Corresponding authors.}

\author[4]{\fnm{Le} \sur{Zhang}}
\equalcont{Corresponding authors.}

\author[3]{\fnm{Ajad} \sur{Chhatkuli}}

\author[3]{\fnm{Luc Van} \sur{Gool}}

\affil[1]{\orgdiv{Institute for Infocomm Research (I2R)}, \orgname{A*STAR}, \orgaddress{\city{Singapore} \postcode{138632}}}

\affil[2]{\orgdiv{Institute of High Performance Computing (IHPC)}, \orgname{A*STAR}, \orgaddress{\city{Singapore} \postcode{138632}}}

\affil[3]{\orgdiv{Computer Vision Lab}, \orgname{ETH Z\"urich}, \orgaddress{\city{Z\"urich} \postcode{8092}, \country{Switzerland}}}

\affil[4]{\orgdiv{School of Information and Communication Engineering}, \orgname{UESTC}, \orgaddress{\city{Chengdu} \postcode{611731}, \country{China}}}


\abstract{This paper tackles the high computational/space complexity associated with Multi-Head Self-Attention (MHSA) in vanilla vision transformers. To this end, we propose Hierarchical MHSA (H-MHSA), a novel approach that computes self-attention in a hierarchical fashion. Specifically, we first divide the input image into patches as commonly done, and each patch is viewed as a token. Then, the proposed H-MHSA learns token relationships within local patches, serving as local relationship modeling. Then, the small patches are merged into larger ones, and H-MHSA models the global dependencies for the small number of the merged tokens. At last, the local and global attentive features are aggregated to obtain features with powerful representation capacity. Since we only calculate attention for a limited number of tokens at each step, the computational load is reduced dramatically. Hence, H-MHSA can efficiently model global relationships among tokens without sacrificing fine-grained information. With the H-MHSA module incorporated, we build a family of Hierarchical-Attention-based Transformer Networks, namely HAT-Net. To demonstrate the superiority of HAT-Net in scene understanding, we conduct extensive experiments on fundamental vision tasks, including image classification, semantic segmentation, object detection, and instance segmentation. Therefore, HAT-Net provides a new perspective for vision transformers. Code and pretrained models are available at \url{https://github.com/yun-liu/HAT-Net}.}




\maketitle

\section{Introduction}
In the last decade, convolutional neural networks (CNNs) have been the go-to architecture in computer vision, owing to their powerful capability in learning representations from images/videos \cite{krizhevsky2012imagenet,simonyan2015very,he2016deep,ren2016faster,he2017mask,zhao2017pyramid,liu2019richer,liu2022semantic,liu2021dna,liu2020lightweight,liu2020rethinking,liu2020leveraging}.
Meanwhile, in another field of natural language processing (NLP), the transformer architecture \cite{vaswani2017attention} has been the de-facto standard to handle long-range dependencies \cite{devlin2019bert,dai2019transformer}.
Transformers rely heavily on self-attention to model global relationships of sequence data.
Although global modeling is also essential for vision tasks, the 2D/3D structures of vision data make it less straightforward to apply transformers therein. 
This predicament was recently broken by Dosovitskiy \etal \cite{dosovitskiy2021image}, by applying a pure transformer to sequences of image patches.

Motivated by \cite{dosovitskiy2021image}, a large amount of literature on vision transformers has emerged to resolve the problems caused by the domain gap between computer vision and NLP \cite{heo2021rethinking,liu2021swin,wang2021pyramid,xu2021co,fan2021multiscale}.
From our point of view, one major problem of vision transformers is that the sequence length of image patches is much longer than that of tokens (words) in an NLP application, thus leading to high computational/space complexity when computing the Multi-Head Self-Attention (\textbf{MHSA}).
Some efforts have been dedicated to resolving this problem.
ToMe \cite{bolya2023token} improves the throughput of existing ViT models \cite{dosovitskiy2021image} by systematically merging similar tokens through the utilization of a general and light-weight matching algorithm.
PVT \cite{wang2021pyramid} and MViT \cite{fan2021multiscale} downsample the feature to compute attention in a reduced length of tokens but at the cost of losing fine-grained details.
Swin Transformer \cite{liu2021swin} computes attention within small windows to model local relationships, and it gradually enlarges the receptive field by shifting windows and stacking more layers. 
From this point of view, Swin Transformer \cite{liu2021swin} may still be suboptimal because it works in a similar manner to CNNs and needs many layers to model long-range dependencies \cite{dosovitskiy2021image}.

Building upon the discussed strengths of downsampling-based transformers \cite{wang2021pyramid,fan2021multiscale} and window-based transformers \cite{liu2021swin}, each with its distinctive merits, we aim to harness their complementary advantages. Downsampling-based transformers excel at directly modeling global dependencies but may sacrifice fine-grained details, while window-based transformers effectively capture local dependencies but may fall short in global dependency modeling. As widely accepted, both global and local information is essential for visual scene understanding. Motivated by this insight, our approach seeks to amalgamate the strengths of both paradigms, enabling the direct modeling of both global and local dependencies.

To achieve this, we introduce the \textbf{Hierarchical Multi-Head Self-Attention (H-MHSA)}, a novel mechanism that enhances the flexibility and efficiency of self-attention computation in transformers. Our methodology begins by segmenting an image into patches, treating each patch akin to a token \cite{dosovitskiy2021image}. Rather than computing attention across all patches, we further organize these patches into small grids, performing attention computation within each grid. This step is instrumental in capturing local relationships and generating more discriminative local representations. Subsequently, we amalgamate these smaller patches into larger ones and treat the merged patches as new tokens, resulting in a substantial reduction in their number. This enables the direct modeling of global dependencies by calculating self-attention for the new tokens. Ultimately, the attentive features from both local and global hierarchies are aggregated to yield potent features with rich granularities. Notably, as the attention calculation at each step is confined to a small number of tokens, our hierarchical strategy mitigates the computational and space complexity of vanilla transformers. Empirical observations underscore the efficacy of this hierarchical self-attention mechanism, revealing improved generalization results in our experiments.

By simply incorporating H-MHSA, we build a family of \textbf{Hierarchical-Attention-based Transformer Networks (HAT-Net)}. 
To evaluate the efficacy of HAT-Net in scene understanding, we experiment HAT-Net for fundamental vision tasks, including image classification, semantic segmentation, object detection, and instance segmentation. 
Experimental results demonstrate that HAT-Net performs favorably against previous backbone networks.
Note that H-MHSA is based on a very simple and intuitive idea, so H-MHSA is expected to provide a new perspective for the future design of vision transformers.

\section{Related Work}
\noindent\textbf{Convolutional neural networks.}\quad
More than two decades ago, LeCun \etal \cite{lecun1998gradient} built the first deep CNN, \ie, LeNet, for document recognition.
About ten years ago, AlexNet \cite{krizhevsky2012imagenet} introduced pooling layers into CNNs and pushed forward the state of the art of ImageNet classification \cite{russakovsky2015imagenet} significantly.
Since then, CNNs have become the de-facto standard of computer vision owing to its powerful ability in representation learning.
Brilliant achievements have been seen in this direction.
VGGNet \cite{simonyan2015very} investigates networks of increasing depth using small ($3\times 3$) convolution filters.
ResNet \cite{he2016deep} manages to build very deep networks by resolving the gradient vanishing/exploding problem with residual connections \cite{srivastava2015highway}.
GoogLeNet \cite{szegedy2015going} presents the inception architecture \cite{szegedy2016rethinking,szegedy2017inception} using multiple branches with different convolution kernels.
ResNeXt \cite{xie2017aggregated} improves ResNet \cite{he2016deep} by replacing the $3\times 3$ convolution in the bottleneck with a grouped convolution.
DenseNets \cite{huang2017densely} present dense connections, \ie, using the feature maps of all preceding layers as inputs for each layer.
MobileNets \cite{howard2017mobilenets,sandler2018mobilenetv2} decompose the traditional convolution into a pointwise convolution and a depthwise separable convolution for acceleration, and an inverted bottleneck is proposed for ensuring accuracy.
ShuffleNets \cite{zhang2018shufflenet,ma2018shufflenet} further decompose the pointwise convolution into pointwise group convolution and channel shuffle to reduce computational cost.
MansNet \cite{tan2019mnasnet} proposes an automated mobile neural architecture search approach to search for a model with a good trade-off between accuracy and latency.
EfficientNet \cite{tan2019efficientnet} introduces a scaling method to uniformly scale depth/width/resolution dimensions of the architecture searched by MansNet \cite{tan2019mnasnet}.
The above advanced techniques are the engines driving the development of computer vision in the last decade.
This paper aims at improving feature representation learning by designing new transformers.

\myPara{Self-attention mechanism.}
Inspired by the human visual system, the self-attention mechanism is usually adopted to enhance essential information and suppress noisy information.
STN \cite{jaderberg2015spatial} presents the spatial attention mechanism through learning an appropriate spatial transformation for each input.
Chen \etal \cite{chen2017sca} proposed the channel attention model and achieved promising results on the image captioning task.
Wang \etal \cite{wang2017residual} explored self-attention in 
well-known residual networks \cite{he2016deep}.
SENet \cite{hu2020squeeze} applies channel attention to backbone network design and boosts the accuracy of ImageNet classification \cite{russakovsky2015imagenet}.
CBAM \cite{woo2018cbam} sequentially applies channel and spatial attention for adaptive feature refinement in deep networks.
BAM \cite{parkbam} produces a 3D attention map by combining channel and spatial attention.
SK-Net \cite{li2019selective} uses channel attention to selectively fuse multiple branches with different kernel sizes.
Non-local network \cite{wang2018non} presents non-local attention for capturing long-range dependencies.
ResNeSt \cite{zhang2020resnest} is a milestone in this direction.
It applies channel attention on different network branches to capture cross-feature interactions and learn diverse representations.
Our work shares some similarities with these works by applying self-attention for learning feature representations.
The difference is that we propose H-MHSA to learn global relationships rather than a simple feature recalibration using spatial or channel attention in these works.

\myPara{Vision transformer.}
Transformer \cite{vaswani2017attention} entirely relies on self-attention to handle long-range dependencies of sequence data.
It was first proposed for NLP tasks \cite{devlin2019bert,dai2019transformer}.
In order to apply transformers on image data, Dosovitskiy \etal \cite{dosovitskiy2021image} split an image into patches and treated them as tokens.
Then, a pure transformer \cite{vaswani2017attention} can be adopted.
Such a vision transformer (ViT) attains competitive accuracy for ImageNet classification \cite{russakovsky2015imagenet}.
More recently, lots of efforts have been dedicated to improving ViT.
T2T-ViT \cite{yuan2021tokens} proposes to split an image into tokens of overlapping patches so as to represent local structures by surrounding tokens.
CaiT \cite{touvron2021going} builds a deeper transformer network by introducing a per-channel weighting and specific class attention.
DeepViT \cite{zhou2021deepvit} proposes Re-attention to re-generate attention maps to increase their diversity at different layers.
DeiT \cite{touvron2021training} presents a knowledge distillation strategy for improving the training of ViT \cite{dosovitskiy2021image}.
Srinivas \etal \cite{srinivas2021bottleneck} tried to add the bottleneck structure to vision transformers.
Some works build pyramid transformer networks to generate multi-scale features \cite{heo2021rethinking,liu2021swin,wang2021pyramid,xu2021co,fan2021multiscale}.
PVT \cite{wang2021pyramid} adopts convolution operation to downsample the feature map in order to reduce the sequence length in MHSA, thus reducing the computational load.
Similar to PVT \cite{wang2021pyramid}, MViT \cite{fan2021multiscale} utilizes pooling to compute attention on a reduced sequence length.
Swin Transformer \cite{liu2021swin} computes attention within small windows and shifts windows to gradually enlarge the receptive field.
CoaT \cite{xu2021co} computes attention in the channel dimension rather than in the traditional spatial dimension.
ToMe \cite{bolya2023token} enhances the throughput of existing ViT models \cite{dosovitskiy2021image} without requiring retraining, which is achieved by gradually combining similar tokens in a transformer using a matching algorithm.
In this paper, we introduce a novel design to reduce the computational complexity of MHSA and learn both the global and local relationship modeling through vision transformers.

\myPara{Vision MLP networks.}
While CNNs and vision transformers have been widely adopted for computer vision tasks, Tolstikhin \etal \cite{tolstikhin2021mlp} challenged the necessity of convolutions and attention mechanisms. They introduced the MLP-Mixer architecture, which relies solely on multi-layer perceptrons (MLPs). MLP-Mixer incorporates two types of layers: one applies MLPs independently to image patches, facilitating the mixing of per-location features, and the other applies MLPs across patches, enabling the mixing of spatial information. Despite lacking convolutions and attention, MLP-Mixer demonstrated competitive performance in image classification compared to state-of-the-art models. Liu \etal \cite{liu2021pay} introduced gMLP, an MLP-based model with gating, showcasing its comparable performance to transformers in crucial language and vision applications. In contrast to other MLP-like models that encode spatial information along flattened spatial dimensions, Vision Permutator \cite{hou2022vision} uniquely encodes feature representations along height and width dimensions using linear projections. Wang \etal \cite{wang2022parameterization} proposed a novel positional spatial gating unit, leveraging classical relative positional encoding to efficiently capture cross-token relations for token mixing. Despite these advancements, the performance of vision MLP networks still lags behind that of vision transformers. In this paper, we focus on the design of a new vision transformer network.

\begin{figure*}[!t]
    \centering
    \includegraphics[width=.85\textwidth]{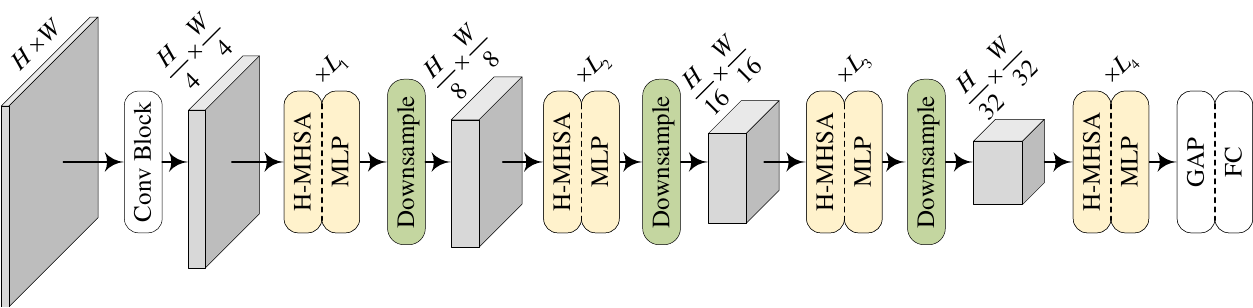}
    \caption{\textbf{Illustration of the proposed HAT-Net.} GAP: global average pooling; FC: fully-connected layer. $\times L_i$ means that the transformer block is repeated for $L_i$ times. $H$ and $W$ denote the height and width of the input image, respectively.}
    \label{fig:net}
\end{figure*}

\section{Methodology}
In this section, we first provide a brief review of vision transformers \cite{dosovitskiy2021image} in \secref{sec:review}.
Then, we present the proposed H-MHSA and analyze its computational complexity in \secref{sec:attn}.
Finally, we describe the configuration details of the proposed HAT-Net in \secref{sec:net}.

\subsection{Review of Vision Transformers}\label{sec:review}
Transformer \cite{vaswani2017attention,dosovitskiy2021image} heavily relies on MHSA to model long-range relationships.
Suppose $\mathbf{X} \in \mathbb{R}^{H\times W\times C}$ denotes the input, where $H$, $W$, and $C$ are the height, width, and the feature dimension, respectively.
We reshape $\mathbf{X}$ and define the query $\mathbf{Q}$, key $\mathbf{K}$, value $\mathbf{V}$ as
\begin{equation}
\begin{aligned}
\mathbf{X} \in \mathbb{R}^{H\times W\times C} &\to \mathbf{X} \in \mathbb{R}^{(H\times W)\times C}, \\
\mathbf{Q} = \mathbf{X}\mathbf{W}^q, \qquad
\mathbf{K} &= \mathbf{X}\mathbf{W}^k, \qquad
\mathbf{V} = \mathbf{X}\mathbf{W}^v,
\end{aligned}
\end{equation}
where $\mathbf{W}^q \in \mathbb{R}^{C\times C}$, $\mathbf{W}^k \in \mathbb{R}^{C\times C}$, and $\mathbf{W}^v \in \mathbb{R}^{C\times C}$ are the trainable weight matrices of linear transformations.
With a mild assumption that the input and output have the same dimension, the traditional MHSA can be formulated as 
\begin{equation}\label{equ:mhsa}
\mathbf{A} = {\rm Softmax}(\mathbf{Q}\mathbf{K}^{\rm T} / \sqrt{d}) \mathbf{V},
\end{equation}
in which $\sqrt{d}$ means an approximate normalization, and the ${\rm Softmax}$ function is applied to the rows of the matrix.
Note that we omit the concept of multiple heads here for simplicity.
In \equref{equ:mhsa}, the matrix product of $\mathbf{Q}\mathbf{K}^{\rm T}$ first computes the similarity between each pair of tokens.
Each new token is then derived over the combination of all tokens according to the similarity.
After the computation of MHSA, a residual connection is further added to ease the optimization, like
\begin{equation}\label{equ:residual}
\begin{aligned}
\mathbf{X} \in \mathbb{R}^{(H\times W)\times C} &\to \mathbf{X} \in \mathbb{R}^{H\times W\times C}, \\
\mathbf{A'} &= \mathbf{A}\mathbf{W}^p + \mathbf{X},
\end{aligned}
\end{equation}
in which $\mathbf{W}^p\in \mathbb{R}^{C\times C}$ is a trainable weight matrix for feature projection.
At last, a multilayer perceptron (MLP) is adopted to enhance the representation, like
\begin{equation}\label{equ:mlp}
\mathbf{Y} = {\rm MLP}(\mathbf{A'}) + \mathbf{A'},
\end{equation}
where $\mathbf{Y}$ denotes the output of a transformer block.

It is easy to infer that the computational complexity of MHSA (\equref{equ:mhsa}) is
\begin{equation}\label{equ:complexity}
\Omega({\rm MHSA}) = 3HWC^2 + 2H^2 W^2 C.
\end{equation}
Similarly, the space complexity (memory consumption) also includes the term of $O(H^2 W^2)$. 
As commonly known, $O(H^2 W^2)$ could become very large for high-resolution inputs.
This limits the applicability of transformers for vision tasks. Motivated by this, we aim at improving MHSA to reduce such complexity and maintain the capacity of global relationship modeling without the risk of sacrificing performances.

\newcommand{\splitcell}[1]{\begin{tabular}{@{}c@{}}#1\end{tabular}}
\newcommand{\bsplitcell}[1]{$\left[\splitcell{#1}\right]$}

\begin{table*}[!tb]
\centering
\renewcommand{\tabcolsep}{4.mm}
\caption{\textbf{Network configurations of HAT-Net.} The settings of building blocks are shown in brackets, with the number of blocks stacked. For the first stage, each convolution has $C$ channels and a stride of $S$. For the other four stages, each MLP uses a $K\times K$ DW-Conv and an expansion ratio of $E$. Note that we omit the downsampling operation after the $t$-th stage ($t=\{2,3,4\}$) for simplicity. ``\#Param'' refers to the number of parameters.}
\label{tab:networks}
\resizebox{\linewidth}{!}{%
\begin{tabular}{c|c|c|c|c|c|c} \toprule
    Stage & Input Size & Operator & HAT-Net-Tiny & HAT-Net-Small & HAT-Net-Medium & HAT-Net-Large
    \\ \midrule
    1 & $224\times 224$ & $3\times 3$ conv.
    & \splitcell{$C=16$, $S=2$\\ $C=48$, $S=2$}
    & \splitcell{$C=16$, $S=2$\\ $C=64$, $S=2$}
    & \splitcell{$C=16$, $S=2$\\ $C=64$, $S=2$}
    & \splitcell{$C=16$, $S=2$\\ $C=64$, $S=2$}
    \\ \midrule
    2 & $56\times 56$ & \splitcell{H-MHSA \\ MLP}
    & \bsplitcell{$C=48$\\ $K=3$\\ $E=8$} $\times$ 2
    & \bsplitcell{$C=64$\\ $K=3$\\ $E=8$} $\times$ 2
    & \bsplitcell{$C=64$\\ $K=5$\\ $E=8$} $\times$ 3
    & \bsplitcell{$C=64$\\ $K=3$\\ $E=8$} $\times$ 3
    \\ \midrule
    3 & $28\times 28$ & \splitcell{H-MHSA \\ MLP}
    & \bsplitcell{$C=96$\\  $K=3$\\ $E=8$} $\times$ 2
    & \bsplitcell{$C=128$\\ $K=3$\\ $E=8$} $\times$ 3
    & \bsplitcell{$C=128$\\ $K=3$\\ $E=8$} $\times$ 6
    & \bsplitcell{$C=128$\\ $K=3$\\ $E=8$} $\times$ 8
    \\ \midrule
    4 & $14\times 14$ & \splitcell{H-MHSA \\ MLP}
    & \bsplitcell{$C=240$\\ $K=3$\\ $E=4$} $\times$ 6
    & \bsplitcell{$C=320$\\ $K=3$\\ $E=4$} $\times$ 8
    & \bsplitcell{$C=320$\\ $K=5$\\ $E=4$} $\times$ 18
    & \bsplitcell{$C=320$\\ $K=3$\\ $E=4$} $\times$ 27
    \\ \midrule
    5 & $7\times 7$ & \splitcell{H-MHSA \\ MLP}
    & \bsplitcell{$C=384$\\ $K=3$\\ $E=4$} $\times$ 3
    & \bsplitcell{$C=512$\\ $K=3$\\ $E=4$} $\times$ 3
    & \bsplitcell{$C=512$\\ $K=3$\\ $E=4$} $\times$ 3
    & \bsplitcell{$C=640$\\ $K=3$\\ $E=4$} $\times$ 3
    \\ \midrule
    & $1\times 1$ & - 
    & \multicolumn{4}{c}{Global Average Pooling, 1000-d FC, Softmax}
    \\ \midrule
    \multicolumn{3}{c|}{\#Param} & 12.7M & 25.7M & 42.9M & 63.1M 
    \\ \bottomrule
\end{tabular}}
\end{table*}

\subsection{Hierarchical Multi-Head Self-Attention}
\label{sec:attn}
In this section, we present an approach to alleviate the computational and space demands associated with \equref{equ:mhsa} through the utilization of our proposed H-MHSA mechanism. Rather than computing attention over the entire input, we adopt a hierarchical strategy, allowing each step to process only a limited number of tokens.

The initial step concentrates on local attention computation. Assuming the input feature map is denoted as $\mathbf{X} \in \mathbb{R}^{H \times W \times C}$, we partition the feature map into small grids of size $G_1\times G_1$ and reshape it as follows:
\begin{equation}\label{equ:local_reshape}
\begin{aligned}
\mathbf{X} \in \mathbb{R}^{H \times W \times C} 
& \to \mathbf{X}_1 \in \mathbb{R}^{(\frac{H}{G_1} \times G_1) \times (\frac{W}{G_1} \times G_1) \times C} \\
&\to \mathbf{X}_1 \in \mathbb{R}^{(\frac{H}{G_1} \times \frac{W}{G_1}) \times (G_1 \times G_1) \times C}.
\end{aligned}
\end{equation}
The query, key, and value are then calculated by 
\begin{equation}
\mathbf{Q}_1 = \mathbf{X}_1 \mathbf{W}^q_1, \quad
\mathbf{K}_1 = \mathbf{X}_1 \mathbf{W}^k_1, \quad
\mathbf{V}_1 = \mathbf{X}_1 \mathbf{W}^v_1,
\end{equation}
where $\mathbf{W}^q_1, \mathbf{W}^k_1, \mathbf{W}^v_1 \in \mathbb{R}^{C\times C}$ are trainable weight matrices.
Subsequently, \equref{equ:mhsa} is applied to generate the local attentive feature $\mathbf{A}_1$.
To ease network optimization, we reshape $\mathbf{A}_1$ back to the shape of $\mathbf{X}$ through
\begin{equation}
\begin{aligned}
&\mathbf{A}_1 \in \mathbb{R}^{(\frac{H}{G_1} \times \frac{W}{G_1}) \times (G_1 \times G_1) \times C} \\
\to &\mathbf{A}_1 \in \mathbb{R}^{(\frac{H}{G_1} \times G_1) \times (\frac{W}{G_1} \times G_1) \times C} \\
\to &\mathbf{A}_1 \in \mathbb{R}^{H \times W \times C},
\end{aligned}
\end{equation}
and incorporate a residual connection:
\begin{equation} \label{equ:local_residual}
\mathbf{A}_1 = \mathbf{A}_1 + \mathbf{X}.
\end{equation}
As the local attentive feature $\mathbf{A}_1$ is computed within each small $G_1 \times G_1$ grid, a substantial reduction in computational and space complexity is achieved.

The second step focuses on global attention calculation. Here, we downsample $\mathbf{A}_1$ by a factor of $G_2$ during the computation of key and value matrices. This downsampling enables efficient global attention calculation, treating each $G_2 \times G_2$ grid as a token. This process can be expressed as
\begin{equation} \label{equ:global_ds}
\widehat{\mathbf{A}}_1 = {\rm AvePool}_{G_2}(\mathbf{A}_1),
\end{equation}
where ${\rm AvePool}_{G_2}(\cdot)$ denotes downsampling a feature map by $G_2$ times using average pooling with both the kernel size and stride set to $G_2$. 
Consequently, we have $\widehat{\mathbf{A}}_1 \in \mathbb{R}^{\frac{H}{G_2} \times \frac{W}{G_2} \times C}$.
We then reshape $\mathbf{A}_1$ and $\widehat{\mathbf{A}}_1$ as follows: 
\begin{equation}\label{equ:step_i}
\begin{aligned}
\mathbf{A}_1 \in \mathbb{R}^{H \times W \times C} &\to \mathbf{A}_1 \in \mathbb{R}^{(H \times W) \times C}, \\
\widehat{\mathbf{A}}_1 \in \mathbb{R}^{\frac{H}{G_2} \times \frac{W}{G_2} \times C} &\to \widehat{\mathbf{A}}_1 \in \mathbb{R}^{(\frac{H}{G_2} \times \frac{W}{G_2}) \times C}.
\end{aligned}
\end{equation}
Following this, we compute the query, key, and value as
\begin{equation}
\mathbf{Q}_2 = \mathbf{A}_1 \mathbf{W}^q_2, \quad
\mathbf{K}_2 = \widehat{\mathbf{A}}_1 \mathbf{W}^k_2, \quad
\mathbf{V}_2 = \widehat{\mathbf{A}}_1 \mathbf{W}^v_2,
\end{equation}
where $\mathbf{W}^q_2, \mathbf{W}^k_2, \mathbf{W}^v_2 \in \mathbb{R}^{C\times C}$ are trainable weight matrices.
It is easy to derive that we have $\mathbf{Q}_2 \in \mathbb{R}^{(H \times W) \times C}$, $\mathbf{K}_2 \in \mathbb{R}^{(\frac{H}{G_2} \times \frac{W}{G_2}) \times C}$, and $\mathbf{V}_2 \in \mathbb{R}^{(\frac{H}{G_2} \times \frac{W}{G_2}) \times C}$.
Subsequently, \equref{equ:mhsa} is called to obtain the global attentive feature $\mathbf{A}_2 \in \mathbb{R}^{(H \times W) \times C}$, followed by a reshaping operation:
\begin{equation} \label{equ:global_out}
\mathbf{A}_2 \in \mathbb{R}^{(H \times W) \times C} \to \mathbf{A}_2 \in \mathbb{R}^{H \times W \times C}.
\end{equation}
The final output of H-MHSA is given by
\begin{equation}\label{equ:output}
\text{H-MHSA}(\mathbf{X}) = (\mathbf{A}_1 + \mathbf{A}_2) \mathbf{W}^p + \mathbf{X},
\end{equation}
where $\mathbf{W}^p$ has the same meaning as in \equref{equ:residual}.
In this way, H-MHSA effectively models both local and global relationships, akin to vanilla MHSA.

The computational complexity of H-MHSA can be expressed as
\begin{equation} \small
\Omega(\text{H-MHSA}) = HWC(4C + 2G_1^2) + 2\frac{HW}{G_2^2}C (C + HW).
\end{equation}
Compared to \equref{equ:complexity}, this represents a reduction in computational complexity from $O(H^2W^2)$ to $O(HW G_1^2 + \frac{H^2 W^2}{G_2^2})$. The same conclusion can be easily derived for space complexity.

We continue by comparing H-MHSA with existing vision transformers, highlighting distinctive features. Swin Transformer \cite{liu2021swin} focuses on modeling local relationships, progressively expanding the receptive field through shifted windows and additional layers. Conversely, PVT \cite{wang2021pyramid} prioritizes global relationships through downsampling key and value matrices but overlooks local information. In contrast, our proposed H-MHSA excels by concurrently capturing both local and global relationships. While Swin Transformer employs a fixed window size (\ie, a fixed-size bias matrix), and PVT uses a constant downsampling ratio (\ie, a convolution with the kernel size equal to the stride), these approaches necessitate retraining on the ImageNet dataset \cite{russakovsky2015imagenet} for any re-parameterization. In contrast, the parameter-free nature of $G_1$ and $G_2$ in H-MHSA allows flexible configuration adjustments for downstream vision tasks without the need for retraining on ImageNet.

In computer vision, achieving a comprehensive understanding of scenes relies on the simultaneous consideration of both global and local information. Within the framework of our proposed H-MHSA, global self-attention calculation (\equref{equ:global_ds}-(\ref{equ:global_out})) is instrumental in establishing the foundation for scene interpretation, enabling the recognition of overarching patterns and aiding in high-level decision-making processes. Concurrently, local self-attention calculation (\equref{equ:local_reshape}-(\ref{equ:local_residual})) is crucial for refining the understanding of individual components within the larger context, facilitating more detailed and nuanced scene analysis. H-MHSA excels in striking the delicate balance between global and local information, thereby facilitating a nuanced and accurate comprehension of diverse scenes. In essence, the seamless integration of global and local self-attention within the H-MHSA framework empowers transformers to navigate the intricacies of scene understanding, facilitating context-aware decision-making.

\subsection{Network Architecture} \label{sec:net}
This part introduces the network architecture of HAT-Net.
We follow the common practice in CNNs to use a global average pooling layer and a fully connected layer to predict image classes \cite{liu2021swin}. 
This is different from existing transformers which rely on another 1D class token to make predictions \cite{dosovitskiy2021image,heo2021rethinking,wang2021pyramid,xu2021co,fan2021multiscale,yuan2021tokens,touvron2021going,zhou2021deepvit,touvron2021training,han2021transformer,li2021localvit,yuan2021incorporating}.
We also observe that existing transformers \cite{dosovitskiy2021image,heo2021rethinking,liu2021swin,wang2021pyramid,xu2021co,fan2021multiscale,yuan2021tokens,touvron2021going,zhou2021deepvit,touvron2021training} usually adopt the GELU function \cite{hendrycks2016gaussian} for nonlinear activation. 
However, GELU is memory-hungry during network training. 
We empirically found that the SiLU function \cite{elfwing2018sigmoid}, originally coined in \cite{hendrycks2016gaussian}, performs on-par with GELU and is more memory-friendly.
Hence, HAT-Net uses SiLU \cite{elfwing2018sigmoid} for nonlinear activation.
Besides, we add a depthwise separable convolution (DW-Conv) \cite{howard2017mobilenets} inside the MLP as widely done.

The overall architecture of HAT-Net is illustrated in \figref{fig:net}.
At the beginning of HAT-Net, instead of flattening image patches \cite{dosovitskiy2021image}, we apply two sequential vanilla $3\times 3$ convolutions, each of which has a stride of 2, to downsample the input image into $1/4$ scale. 
Then, we stack H-MHSA and MLP alternatively, which can be divided into four stages with pyramid feature scales of $1/4$, $1/8$, $1/16$, and $1/32$, respectively.
For feature downsampling at the end of each stage, a vanilla $3\times 3$ convolution with a stride of 2 is used.
The configuration details of HAT-Net are summarized in \tabref{tab:networks}.
We provide four versions of HAT-Net: HAT-Net-Tiny, HAT-Net-Small, HAT-Net-Medium, and HAT-Net-Large, whose number of parameters is similar to ResNet18, ResNet50, ResNet101, and ResNet152 \cite{he2016deep}, respectively.
We only adopt simple parameter settings without careful tuning to demonstrate the effectiveness and generality of HAT-Net.
The dimension of each head in the multi-head setting is set to 48 for HAT-Net-Tiny and 64 for other versions.

To enhance the applicability of HAT-Net across diverse vision tasks, we present guidelines for configuring the parameter-free $G_1$ and $G_2$. While established models like Swin Transformer \cite{liu2021swin} adhere to a fixed window size of 7, and PVT \cite{wang2021pyramid} employs a set of constant downsampling ratios ${8,4,2}$ for the $t$-th stage ($t={2,3,4}$), we advocate for certain adjustments. Practically, we find that a window size of 8 is more pragmatic than 7, given that input resolutions often align with multiples of 8. Moreover, augmenting downsampling ratios serves to mitigate computational complexity. Consequently, for image classification on the ImageNet dataset \cite{russakovsky2015imagenet}, where the standard input resolution is $224\times 224$ pixels, we designate $G_1={8,7,7}$ and $G_2={8,4,2}$ for the $t$-th stage ($t={2,3,4}$). Here, a window size of 7 is necessitated by the chosen resolution, and small downsampling rates are in line with the approach taken by PVT \cite{wang2021pyramid}. In scenarios involving downstream tasks like semantic segmentation, object detection, and instance segmentation, where input resolutions tend to be larger, we opt for $G_1={8,8,8}$ for convenience and $G_2={16,8,4}$ to curtail computational expenses. For a comprehensive analysis of the impact of different $G_1$ and $G_2$ settings, we conduct an ablation study in \secref{sec:ablation}.

\section{Experiments}
To show the superiority of HAT-Net in feature representation learning, this section evaluates HAT-Net for image classification, semantic segmentation, object detection and instance segmentation.

\subsection{Image Classification}
\textbf{Experimental setup.}\quad
The ImageNet dataset \cite{russakovsky2015imagenet} consists of 1.28M training images and 50K validation images from 1000 categories.
We adopt the training set to train our networks and the validation set to test the performance.
We implement HAT-Net using the popular PyTorch framework \cite{paszke2019pytorch}.
For a fair comparison, we follow the same training protocol 
as DeiT \cite{touvron2021training}, which is the standard protocol for training transformer networks nowadays.
Specifically, the input images are randomly cropped to $224\times 224$ pixels, followed by random horizontal flipping and \textit{mixup} \cite{zhang2018mixup} for data augmentation.
Label smoothing \cite{szegedy2016rethinking} is used to avoid overfitting.
The AdamW optimizer \cite{loshchilov2019decoupled} is adopted with the momentum of 0.9, the weight decay of 0.05, and a mini-batch size of 128 per GPU by default.
The initial learning rate is set to 1$e$-3, which decreases following the cosine learning rate schedule \cite{loshchilov2016sgdr}.
The training process lasts for 300 epochs on eight NVIDIA Tesla V100 GPUs.
Note that for ablation studies, we utilize a mini-batch size
of 64 and 100 training epochs to save training time.
Moreover, we set $G_1=\{8,7,7\}$ and $G_2=\{8,4,2\}$ for $t$-th stage ($t=\{2,3,4\}$), respectively.
The fifth stage can be processed directly using the vanilla MHSA mechanism.
For model evaluation, we apply a center crop of $224\times 224$ pixels on validation images to evaluate the recognition accuracy.
We report the top-1 classification accuracy on the ImageNet validation set \cite{russakovsky2015imagenet} as well as the number of parameters and the number of FLOPs for each model.

\def\M{\text{M}}
\def\G{\text{G}}

\begin{table}[!t]
\centering
\renewcommand{\tabcolsep}{1.mm}
\caption{\textbf{Comparison to \sArt methods on the ImageNet validation set \cite{russakovsky2015imagenet}.} ``*'' indicates the performance of a method using the default training setting in the original paper. ``\#Param'' and ``\#FLOPs'' refer to the number of parameters and the number of FLOPs, respectively. ``$\dag$'' marks models that use the input size of $384\times 384$; otherwise, models use the input size of $224\times 224$.}
\label{tab:imagenet}
\resizebox{\linewidth}{!}{%
\begin{tabular}{c|l|D{.}{.}{-1}|D{.}{.}{-1}|c} \toprule
    Arch. & \multicolumn{1}{c|}{Models} & \multicolumn{1}{c|}{\#Param} & \multicolumn{1}{c|}{\#FLOPs} & Top-1 Acc. \\ \midrule
    \multirow{2}{*}{CNN} & ResNet18* \cite{he2016deep} & 11.7\M & 1.8\G & 69.8 \\ 
    & ResNet18 \cite{he2016deep} & 11.7\M & 1.8\G & 68.5 \\ \cmidrule{1-1}
    \multirow{4}{*}{Trans} & DeiT-Ti/16 \cite{touvron2021training} & 5.7\M & 1.3\G & 72.2 \\ 
    & PVT-Tiny \cite{wang2021pyramid} & 13.2\M & 1.9\G & 75.1 \\ 
    & PVTv2-B1 \cite{wang2021pvtv2} & 13.1\M & 2.1\G & 78.7 \\ 
    & \cellcolor{mygray}HAT-Net-Tiny & \cellcolor{mygray}12.7\M & \cellcolor{mygray}2.0\G & \cellcolor{mygray}79.8 \\ \midrule
    \multirow{6}{*}{CNN} & ResNet50* \cite{he2016deep} & 25.6\M & 4.1\G & 76.1 \\ 
    & ResNet50 \cite{he2016deep} & 25.6\M & 4.1\G & 78.5 \\
    & ResNeXt50-32x4d* \cite{xie2017aggregated} & 25.0\M & 4.3\G & 77.6 \\
    & ResNeXt50-32x4d \cite{xie2017aggregated} & 25.0\M & 4.3\G & 79.5 \\
    & RegNetY-4G \cite{radosavovic2020designing} & 20.6\M & 4.0\G & 80.0 \\
    & ResNeSt-50 \cite{zhang2020resnest} & 27.5\M & 5.4\G & 81.1 \\ \cmidrule{1-1}
    & ToMe-ViT-S/16 \cite{bolya2023token} & 22.1\M & 2.7\G & 79.4 \\
    \multirow{9}{*}{Trans} & DeiT-S/16 \cite{touvron2021training} & 22.1\M & 4.6\G & 79.8 \\
    & T2T-ViT$_t$-14 \cite{yuan2021tokens} & 21.5\M & 5.2\G & 80.7 \\
    & TNT-S \cite{han2021transformer} & 23.8\M & 5.2\G & 81.3 \\
    & CvT-13 \cite{wu2021cvt} & 20.0\M & 4.5\G & 81.6 \\
    & PVT-Small \cite{wang2021pyramid} & 24.5\M & 3.8\G & 79.8 \\
    & PVTv2-B2 \cite{wang2021pvtv2} & 25.4\M & 4.0\G & 82.0 \\
    & Swin-T \cite{liu2021swin} & 28.3\M & 4.5\G & 81.3 \\
    & Twins-SVT-S \cite{chu2021twins} & 24.0\M & 2.8\G & 81.7 \\
    & \cellcolor{mygray}HAT-Net-Small & \cellcolor{mygray}25.7\M & \cellcolor{mygray}4.3\G & \cellcolor{mygray}82.6 \\ \midrule
    \multirow{6}{*}{CNN} & ResNet101* \cite{he2016deep} & 44.7\M & 7.9\G & 77.4 \\
    & ResNet101 \cite{he2016deep} & 44.7\M & 7.9\G & 79.8 \\
    & ResNeXt101-32x4d* \cite{xie2017aggregated} & 44.2\M & 8.0\G & 78.8 \\
    & ResNeXt101-32x4d \cite{xie2017aggregated} & 44.2\M & 8.0\G & 80.6 \\
    & RegNetY-8G \cite{radosavovic2020designing} & 39.2\M & 8.0\G & 81.7 \\
    & ResNeSt-101 \cite{zhang2020resnest} & 48.3\M & 10.3\G & 83.0 \\ \cmidrule{1-1}
    \multirow{6}{*}{Trans} & T2T-ViT$_t$-19 \cite{yuan2021tokens} & 39.2\M & 8.4\G & 81.4 \\
    & CvT-21 \cite{wu2021cvt} & 31.5\M & 7.1\G & 82.5 \\
    & MViT-B-16 \cite{fan2021multiscale} & 37.0\M & 7.8\G & 82.5 \\
    & PVT-Medium \cite{wang2021pyramid} & 44.2\M & 6.7\G & 81.2 \\
    & PVTv2-B3 \cite{wang2021pvtv2} & 45.2\M & 6.9\G & 83.2 \\
    & Swin-S \cite{liu2021swin} & 49.6\M & 8.7\G & 83.0 \\
    & \cellcolor{mygray}HAT-Net-Medium & \cellcolor{mygray}42.9\M & \cellcolor{mygray}8.3\G & \cellcolor{mygray}84.0 \\ \midrule
    \multirow{3}{*}{CNN} & ResNet152* \cite{he2016deep} & 60.2\M & 11.6\G & 78.3 \\
    & ResNeXt101-64x4d* \cite{xie2017aggregated} & 83.5\M & 15.6\G & 79.6 \\
    & ResNeXt101-64x4d \cite{xie2017aggregated} & 83.5\M & 15.6\G & 81.5 \\ \cmidrule{1-1}
    \multirow{9}{*}{Trans} & ViT-B/16$^\dag$ \cite{dosovitskiy2021image} & 86.6\M & 55.4\G & 77.9 \\
    & ViT-L/16$^\dag$ \cite{dosovitskiy2021image} & 304.3\M & 190.7\G & 76.5 \\
    & ToMe-ViT-L/16 \cite{bolya2023token} & 304.3\M & 22.3\G & 84.2 \\
    & DeiT-B/16 \cite{touvron2021training} & 86.6\M & 17.6\G & 81.8 \\
    & MViT-B-24 \cite{fan2021multiscale} & 53.5\M & 10.9\G & 83.1 \\
    & TNT-B \cite{han2021transformer} & 65.6\M & 14.1\G & 82.8 \\
    & PVT-Large \cite{wang2021pyramid} & 61.4\M & 9.8\G & 81.7 \\
    & PVTv2-B4 \cite{wang2021pvtv2} & 62.6\M & 10.1\G & 83.6 \\
    & Swin-B \cite{liu2021swin} & 87.8\M & 15.4\G & 83.3 \\
    & Twins-SVT-B \cite{chu2021twins} & 56.0\M & 8.3\G & 83.2 \\
    & \cellcolor{mygray}HAT-Net-Large & \cellcolor{mygray}63.1\M & \cellcolor{mygray}11.5\G & \cellcolor{mygray}84.2 \\ \bottomrule
\end{tabular}}
\end{table}

\myPara{Experimental results.}
We compare HAT-Net with \sArt network architectures, including CNN-based ones like ResNet \cite{he2016deep}, ResNeXt \cite{xie2017aggregated}, RegNetY \cite{radosavovic2020designing}, ResNeSt \cite{zhang2020resnest}, and transformer-based ones like ViT \cite{dosovitskiy2021image}, DeiT \cite{touvron2021training}, T2T-ViT \cite{yuan2021tokens}, TNT \cite{han2021transformer}, CvT \cite{wu2021cvt}, MViT \cite{fan2021multiscale}, PVT \cite{wang2021pyramid}, Swin Transformer \cite{liu2021swin}, Twins \cite{chu2021twins}, ToMe \cite{bolya2023token}.
The results are summarized in \tabref{tab:imagenet}.
We can observe that HAT-Net achieves \sArt performance.
Specifically, with similar numbers of parameters and FLOPs,
HAT-Net-Tiny, HAT-Net-Small, HAT-Net-Medium, and HAT-Net-Large outperforms the second best results by 1.1\%, 0.6\%, 0.8\%, and 0.6\% in terms of the top-1 accuracy, respectively.
Since the performance for image classification implies the ability of a network for learning feature representations, the above comparison suggests that the proposed HAT-Net has great potential for generic scene understanding.

\begin{table}[!t]
    \centering
    \setlength{\tabcolsep}{1mm}
    \caption{\textbf{Experimental results on the ADE20K validation dataset \cite{zhou2017scene} for semantic segmentation.} We replace the backbone of Semantic FPN \cite{kirillov2019panoptic} with various network architectures. The number of FLOPs is calculated with the input size of $512\times 512$.}
    \label{tab:exp_ade20k}
    \resizebox{\columnwidth}{!}{%
    \begin{tabular}{l|c|c|c} \toprule
        \multirow{2}[0]{*}[-1.5mm]{Backbone} & \multicolumn{3}{c}{Semantic FPN \cite{kirillov2019panoptic}}
        \\ \cmidrule{2-4}
        & \#Param (M) $\downarrow$ & FLOPs (G) $\downarrow$ & \multicolumn{1}{c}{mIoU (\%) $\uparrow$}
        \\ \midrule
        ResNet-18 \cite{he2016deep} & 15.5 & 31.9 & 32.9 \\
        PVT-Tiny \cite{wang2021pyramid} & 17.0 & 32.1 & 35.7 \\
        PVTv2-B1~\cite{wang2021pvtv2} & 17.8 & 33.1 & 41.5 \\
        \rowcolor{mygray} HAT-Net-Tiny & 15.9 & 33.2 & 43.6 \\
        \midrule
        ResNet-50 \cite{he2016deep} & 28.5 & 45.4  & 36.7 \\
        PVT-Small \cite{wang2021pyramid} & 28.2 & 42.9 & 39.8 \\
        Swin-T~\cite{liu2021swin} & 31.9 & 46 & 41.5 \\
        Twins-SVT-S~\cite{chu2021twins} & 28.3  & 37 & 43.2 \\
        PVTv2-B2~\cite{wang2021pvtv2} & 29.1 & 44.1 & 46.1 \\
        \rowcolor{mygray} HAT-Net-Small & 29.5 & 49.6 & 46.6 \\
        \midrule
        ResNet-101 \cite{he2016deep} & 47.5 & 64.8 & 38.8 \\
        ResNeXt-101-32x4d \cite{xie2017aggregated} & 47.1 & 64.6 & 39.7 \\
        PVT-Medium \cite{wang2021pyramid} & 48.0 & 59.4 & 41.6 \\
        Swin-S \cite{liu2021swin} & 53.2 & 70 & 45.2 \\
        Twins-SVT-B \cite{chu2021twins} & 60.4 & 67 & 45.3 \\
        PVTv2-B3 \cite{wang2021pvtv2} & 49.0 & 60.7 & 47.3 \\ 
        \rowcolor{mygray} HAT-Net-Medium & 46.7 & 74.7 & 49.3 \\
        \midrule
        ResNeXt-101-64x4d \cite{xie2017aggregated} & 86.4 & 104.2 & 40.2 \\
        PVT-Large \cite{wang2021pyramid} & 65.1 & 78.0 & 42.1 \\
        Swin-B \cite{liu2021swin} & 91.2 & 107 & 46.0 \\
        Twins-SVT-L \cite{chu2021twins} & 102 & 103.7 & 46.7\\
        PVTv2-B4 \cite{wang2021pvtv2} & 66.3 & 79.6 & 48.6 \\
        \rowcolor{mygray} HAT-Net-Large & 66.8 & 96.4 & 49.5 \\
        \bottomrule
    \end{tabular}}
\end{table}

\begin{table*}
\setlength{\tabcolsep}{1.0mm}
\caption{\textbf{Object detection results with RetinaNet \cite{lin2017focal} and instance segmentation results with Mask R-CNN \cite{he2017mask} on the MS-COCO \texttt{val2017} set \cite{lin2014microsoft}.} ``R'' and ``X'' represent ResNet \cite{he2016deep} and ResNeXt \cite{xie2017aggregated}, respectively. The number of FLOPs is computed with the input size of $800 \times 1280$.}
\label{tab:det_seg}
\resizebox{\textwidth}{!}{%
\begin{tabular}{l||cc|lcc|lcc||cc|lcc|lcc} \toprule
\multirow{3}{*}[-3mm]{Backbone} & \multicolumn{8}{c||}{Object Detection} & \multicolumn{8}{c}{Instance Segmentation} \\ \cmidrule{2-17}
& \multirow{2}{*}[-.8mm]{\tabincell{c}{\#Param\\(M) $\downarrow$}} & \multirow{2}{*}[-.8mm]{\tabincell{c}{\#FLOPs\\(G)$\downarrow$}} & \multicolumn{6}{c||}{RetinaNet \cite{lin2017focal}} & \multirow{2}{*}[-.8mm]{\tabincell{c}{\#Param\\(M) $\downarrow$}} & \multirow{2}{*}[-.8mm]{\tabincell{c}{\#FLOPs\\(G) $\downarrow$}} & \multicolumn{6}{c}{Mask R-CNN \cite{he2017mask}} \\
\cmidrule{4-9} \cmidrule{12-17}
& & & AP & AP$_{50}$ & AP$_{75}$ & AP$_S$ & AP$_M$ & AP$_L$ & & & AP$^{\rm b}$ & AP$_{50}^{\rm b}$ & AP$_{75}^{\rm b}$ & AP$^{\rm m}$ & AP$_{50}^{\rm m}$ & AP$_{75}^{\rm m}$ \\ \midrule
%
R-18~\cite{he2016deep} & 21.3 & 190 & 31.8 & 49.6 & 33.6 & 16.3 & 34.3 & 43.2  &31.2 & 209 & 34.0 & 54.0 & 36.7 & 31.2 & 51.0 & 32.7 \\
ViL-Tiny~\cite{zhang2021multi} & 16.6 & 204 & 40.8 & 61.3 & 43.6 & 26.7 & 44.9 & 53.6 & 26.9 & 223 & 41.4 & 63.5 & 45.0 & 38.1 & 60.3 & 40.8 \\
PVT-Tiny~\cite{wang2021pyramid} & 23.0 & 205 & 36.7 & 56.9 & 38.9 & 22.6 & 38.8 & 50.0 & 32.9 & 223 & 36.7 & 59.2 & 39.3 & 35.1 & 56.7 & 37.3 \\
\rowcolor{mygray} HAT-Net-Tiny & 21.6 & 212 & 42.5 & 63.3 & 45.8 & 26.9 & 46.1 & 56.6 & 31.8 & 231 & 43.1 & 65.4 & 47.4 & 39.7 & 62.5 & 42.4 \\ \midrule
R-50~\cite{he2016deep} & 37.7 & 239 & 36.3 & 55.3 & 38.6 & 19.3 & 40.0 & 48.8 & 44.2& 260 & 38.0 & 58.6 & 41.4 & 34.4 & 55.1 & 36.7 \\
PVT-Small~\cite{wang2021pyramid} & 34.2 & 261 & 40.4 & 61.3 & 43.0 & 25.0 & 42.9 & 55.7 & 44.1 & 280 & 40.4 & 62.9 & 43.8 & 37.8 & 60.1 & 40.3 \\
Swin-T \cite{liu2021swin} & 38.5 & 248 & 41.5 & 62.1 & 44.2 & 25.1 & 44.9 & 55.5 & 47.8 & 264 & 42.2 & 64.6 & 46.2 & 39.1 & 61.6 & 42.0\\
ViL-Small~\cite{zhang2021multi} & 35.7 & 292 & 44.2 & 65.2 & 47.6 & 28.8 & 48.0 & 57.8 & 45.0 & 310 & 44.9 & 67.1 & 49.3 & 41.0 & 64,2 & 44.1 \\
Twins-SVT-S~\cite{chu2021twins} & 34.3 & 236 & 43.0 & 64.2 & 46.3 & 28.0 & 46.4 & 57.5 & 44.0 & 254 & 43.4 & 66.0 & 47.3 & 40.3 & 63.2 & 43.4 \\
\rowcolor{mygray} HAT-Net-Small & 35.5 & 286 & 44.8 & 65.8 & 48.1 & 28.8 & 48.6 & 59.5 & 45.4 & 303 & 45.2 & 67.6 & 49.9 & 41.6 & 64.6 & 44.7 \\ \midrule
R-101~\cite{he2016deep} &56.7 & 315 & 38.5 & 57.8 & 41.2 & 21.4 & 42.6 & 51.1  &63.2 & 336 & 40.4 & 61.1 & 44.2 & 36.4 & 57.7 & 38.8 \\
X-101-32x4d~\cite{xie2017aggregated} &56.4 & 319 & 39.9 & 59.6 & 42.7 & 22.3 & 44.2 & 52.5 & 62.8 & 340 & 41.9 & 62.5 & 45.9 & 37.5 & 59.4 & 40.2 \\
PVT-Medium~\cite{wang2021pyramid} & 53.9 & 349 & 41.9 & 63.1 & 44.3 & 25.0 & 44.9 & 57.6 & 63.9 & 367 & 42.0 & 64.4 & 45.6 & 39.0 & 61.6 & 42.1 \\
Swin-S~\cite{liu2021swin} & 59.8 & 336 & 44.5 & 65.7 & 47.5 & 27.4 & 48.0 & 59.9 & 69.1 & 354 & 44.8  & 66.6 & 48.9 & 40.9 & 63.4 & 44.2 \\
\rowcolor{mygray} HAT-Net-Medium & 52.7 & 405 & 45.9 & 66.9 & 49.2 & 29.7 & 50.0 & 61.6 & 62.6 & 424 & 47.0 & 69.0 & 51.5 & 42.7 & 66.0 & 46.0 \\ \midrule
X-101-64x4d~\cite{xie2017aggregated} & 95.5 & 473 & 41.0 & 60.9 & 44.0 & 23.9 & 45.2 & 54.0 &101.9 & 493 & 42.8 & 63.8 & {47.3} & 38.4 & 60.6 & 41.3 \\
PVT-Large~\cite{wang2021pyramid} & 71.1 & 450 & 42.6 & 63.7 & 45.4 & 25.8 & 46.0 & 58.4 & 81.0 & 469 & 42.9 & 65.0 & 46.6 & 39.5 & 61.9 & 42.5 \\
Twins-SVT-B~\cite{chu2021twins} & 67.0 & 376 & 45.3 & 66.7 & 48.1 & 28.5 & 48.9 & 60.6 & 76.3 & 395 & 45.2 & 67.6 & 49.3 & 41.5 & 64.5 & 44.8 \\
\rowcolor{mygray} HAT-Net-Large & 73.1 & 519 & 46.3 & 67.2 & 49.6 & 30.0 & 50.6 & 62.4 & 82.7 & 537 & 47.4 & 69.3 & 52.1 & 43.1 & 66.5 & 46.6 \\ \bottomrule
\end{tabular}}
\end{table*}

\subsection{Semantic Segmentation}
\textbf{Experimental setup.}\quad
We continue by applying HAT-Net to a fundamental downstream vision task, semantic segmentation, which aims at predicting a class label for each pixel in an image.
We follow \cite{wang2021pyramid,chu2021twins} to replace the backbone of the well-known segmentor, Semantic FPN \cite{kirillov2019panoptic}, with HAT-Net or other backbone networks for a fair comparison.
Experiments are conducted on the challenging ADE20K dataset \cite{zhou2017scene}.
This dataset has 20000 training images, 2000 validation images, and 3302 testing images.
We train Semantic FPN \cite{kirillov2019panoptic} using the training set and evaluate on the validation set.
The training optimizer is AdamW \cite{loshchilov2019decoupled} with weight decay of 1$e$-4.
We apply the \textit{poly} learning rate schedule with $\gamma=0.9$ and the initial learning rate of 1$e$-4.
During training, the batch size is 16, and each image has a resolution of $512\times 512$ through resizing and cropping.
During testing, each image is resized to the shorter side of 512 pixels, without multi-scale testing or flipping.
We adopt the well-known MMSegmentation toolbox \cite{mmseg2020} for the above experiments.
We set $G_1=\{8,8,8\}$ and $G_2=\{16,8,4\}$ for the $t$-th stage ($t=\{2,3,4\}$), respectively.

\myPara{Experimental results.}
The results are depicted in \tabref{tab:exp_ade20k}.
We compare with typical CNN networks, \ie, ResNets \cite{he2016deep} and ResNeXts \cite{xie2017aggregated}, and transformer networks, \ie, Swin Transformer \cite{liu2021swin}, PVT \cite{wang2021pyramid}, PVTv2 \cite{wang2021pvtv2} and Twins-SVT \cite{chu2021twins}.
As can be observed, the proposed HAT-Net achieves significantly better performance than previous competitors.
Specifically, HAT-Net-Tiny, HAT-Net-Small, HAT-Net-Medium, and HAT-Net-Large attain 1.9\%, 0.4\%, 1.9\%, and 0.7\% higher mIoU than the second better results with similar number of parameters and FLOPs.
This demonstrates the superiority of HAT-Net in learning effective feature representations for dense prediction tasks.

\subsection{Object Detection and Instance Segmentation}
\textbf{Experimental setup.}\quad
Since object detection and instance segmentation are also fundamental downstream vision tasks, we apply HAT-Net to both tasks for further evaluating its effectiveness.
Specifically, we utilize two well-known detectors, \ie, RetinaNet \cite{lin2017focal} for object detection and Mask R-CNN \cite{he2017mask} for instance segmentation.
HAT-Net is compared to some well-known CNN and transformer networks by only replacing the backbone of the above two detectors.
Experiments are conducted on the large-scale MS-COCO dataset \cite{lin2014microsoft} by training on the \texttt{train2017} set ($\sim$118K images) and evaluating on the \texttt{val2017} set (5K images).
We adopt MMDection toolbox \cite{chen2019mmdetection} for experiments and follow the experimental settings of PVT \cite{wang2021pyramid} for a fair comparison.
During training, we initialize the backbone weights with the ImageNet-pretrained models.
The detectors are fine-tuned using the AdamW optimizer \cite{loshchilov2019decoupled} with an initial learning rate of 1$e$-4 that is decreased by 10 times after the 8$^\text{th}$ and 11$^\text{th}$ epochs, respectively.
The whole training lasts for 12 epochs with a batch size of 16.
Each image is resized to a shorter side of 800 pixels, but the longer side is not allowed to exceed 1333 pixels.
We set $G_1=\{8,8,8\}$ and $G_2=\{16,8,4\}$ for the $t$-th stage ($t=\{2,3,4\}$), respectively.

\myPara{Experimental results.}
The results are displayed in \tabref{tab:det_seg}.
As can be seen, HAT-Net substantially improves the accuracy over other network architectures with a similar number of parameters.
Twins-SVT \cite{chu2021twins} combines the advantages of PVT \cite{wang2021pyramid} and Swin Transformer \cite{liu2021swin} by alternatively stacking their basic blocks.
When RetinaNet \cite{lin2017focal} is adopted as the detector, HAT-Net-Small attains 1.8\%, 1.6\% and 1.8\% higher results than Twins-SVT-S \cite{chu2021twins} in terms of AP, AP$_\text{50}$ and AP$_\text{75}$, respectively.
Correspondingly, HAT-Net-Large gets 1.0\%, 0.5\% and 1.5\% higher results than Twins-SVT-B \cite{chu2021twins}.
With Mask R-CNN \cite{he2017mask} as the detector, HAT-Net-Large achieves 2.2\%, 1.7\% and 2.8\% higher results than Twins-SVT-B \cite{chu2021twins} in terms of bounding box metrics AP$^\text{b}$, AP$_\text{50}^\text{b}$ and AP$_\text{75}^\text{b}$, respectively.
HAT-Net-Large achieves 1.6\%, 2.0\% and 1.8\% higher results than Twins-SVT-B \cite{chu2021twins} in terms of mask metrics AP$^\text{m}$, AP$_\text{50}^\text{m}$ and AP$_\text{75}^\text{m}$, respectively.
Such significant improvement in object detection and instance 
segmentation shows the superiority of HAT-Net in learning effective
feature representations.

\begin{table}[!t]
\centering
\caption{\textbf{Ablation studies for the hierarchical attention in HAT-Net.} The configuration of HAT-Net-Small is adopted for all experiments. ``\ding{52}'' indicates that we replace the window attention \cite{liu2021swin} with the hierarchical attention at the $i$-th stage. ``Top-1 Acc'' is the top-1 accuracy on the ImageNet validation dataset \cite{russakovsky2015imagenet}. ``mIoU'' is the mean IoU for semantic segmentation on the ADE20K dataset \cite{zhou2017scene}.}
\label{tab:ablation}
\resizebox{\columnwidth}{!}{%
\renewcommand{\tabcolsep}{3.5mm}
\begin{tabular}{c|ccc|cc} \toprule
\multirow{2}{*}{Design} & \multicolumn{3}{c|}{\#Stage} & \multirow{2}{*}{Top-1 Acc} & \multirow{2}{*}{mIoU} \\
& 2 & 3 & 4 & & \\ \midrule
1 &  & & & 78.2 & 42.1\\
2 &   \ding{52} & & & 78.2 & 42.4\\
3 &  \ding{52} & \ding{52} & & 78.4 & 42.5\\
\rowcolor{mygray} 4 & \ding{52} & \ding{52} & \ding{52} & 79.3 & 43.4\\ \bottomrule
\end{tabular}}
\end{table}

\begin{table}[!t]
\centering
\caption{\textbf{Ablation studies for the settings of $G_1$ and $G_2$ in HAT-Net.} The performance assessment is conducted using Mask R-CNN \cite{he2017mask} with HAT-Net-Small as the backbone. Evaluation results are reported on the MS-COCO \texttt{val2017} dataset \cite{lin2014microsoft}.}
\label{tab:ablation_G1G2}
\resizebox{\columnwidth}{!}{%
\renewcommand{\tabcolsep}{.5mm}
\begin{tabular}{c|c|c|ccc|ccc} \toprule
\multicolumn{2}{c|}{Settings} & \multirow{2}{*}[-1.2mm]{\tabincell{c}{\#FLOPs\\(G)$\downarrow$}} & \multicolumn{6}{c}{Mask R-CNN \cite{he2017mask}} \\ \cmidrule{1-2}\cmidrule{4-9}
$G_1$ & $G_2$ & & AP$^{\rm b}$ & AP$_{50}^{\rm b}$ & AP$_{75}^{\rm b}$ & AP$^{\rm m}$ & AP$_{50}^{\rm m}$ & AP$_{75}^{\rm m}$ \\ \midrule
$\{8,8,8\}$ & $\{12,4,2\}$ & 338 & 45.7 & 67.8 & 50.4 & 41.7 & 64.8 & 44.7 \\
$\{8,8,8\}$ & $\{12,6,3\}$ & 313 & 45.3 & 67.6 & 49.7 & 41.6 & 64.8 & 44.9 \\
\rowcolor{mygray} $\{8,8,8\}$ & $\{16,8,4\}$ & 303 & 45.2 & 67.6 & 49.9 & 41.6 & 64.6 & 44.7 \\
$\{8,8,8\}$ & $\{32,16,8\}$ & 291 & 44.6 & 66.8 & 49.1 & 41.0 & 63.8 & 44.3 \\
$\{16,16,8\}$ & $\{16,8,4\}$ & 309 & 45.1 & 67.5 & 49.5 & 41.3 & 64.5 & 44.4 \\
$\{4,4,4\}$ & $\{16,8,4\}$ & 300 & 45.1 & 67.1 & 49.5 & 41.3 & 64.3 & 44.5
\\ \bottomrule
\end{tabular}}
\end{table}

\subsection{Ablation Studies} \label{sec:ablation}
In this part, we evaluate various design choices of the proposed HAT-Net.
As discussed above, we only train all ablation models for 100 epochs to save training time.
The batch size and learning rate are also reduced by half accordingly.
HAT-Net-Small is adopted for these ablation studies.

\myPara{Effect of the proposed H-MHSA.}
Starting from the window attention \cite{liu2021swin} based transformer network, we gradually replace the window attention with our proposed H-MHSA at different stages.
The results are summarized in \tabref{tab:ablation}.
Since the feature map at the fifth stage is small enough for directly computing MHSA, the fifth stage is excluded from \tabref{tab:ablation}.
Note that the first stage of HAT-Net only consists of convolutions so that it is also excluded.
From \tabref{tab:ablation}, we can observe that the performance for both image classification and semantic segmentation is improved when more stages adopt H-MHSA.
This verifies the effectiveness of the proposed H-MHSA in feature presentation learning.
It is interesting to find that the usage of H-MHSA at the fourth stage leads to more significant improvement than other stages.
Intuitively, the fourth stage has the most transformer blocks, so the changes at this stage would lead to more significant effects.

\myPara{A pure transformer version of HAT-Net \vs PVT \cite{wang2021pyramid}.}
When we remove all depthwise separable convolutions from HAT-Net and train the resulting transformer network for 100 epochs, it achieves 77.7\% top-1 accuracy on the ImageNet validation set \cite{russakovsky2015imagenet}.
In contrast, the well-known transformer network, PVT \cite{wang2021pyramid}, attains 75.8\% top-1 accuracy under the same condition.
This suggests that our proposed H-MHSA is very effective in feature representation learning.

\myPara{SiLU \cite{elfwing2018sigmoid} \vs GELU \cite{hendrycks2016gaussian}.}
We use SiLU function \cite{elfwing2018sigmoid}
for nonlinearization rather than the widely-used GELU function
\cite{hendrycks2016gaussian} in transformers 
\cite{vaswani2017attention,dosovitskiy2021image}.
Here, we evaluate the effect of this choice.
HAT-Net with SiLU \cite{elfwing2018sigmoid} attains 82.6\% top-1 accuracy on the ImageNet validation set
\cite{russakovsky2015imagenet} when trained for 300 epochs.
HAT-Net with GELU \cite{hendrycks2016gaussian} gets 82.7\% top-1 accuracy, slightly higher than SiLU \cite{elfwing2018sigmoid}.
However, HAT-Net with GELU \cite{hendrycks2016gaussian} only obtains 45.7\% mIoU on the ADE20K dataset, 0.8\% lower than HAT-Net with SiLU.
When using a batch size of 128 per GPU, HAT-Net with SiLU \cite{elfwing2018sigmoid} occupies 20.2GB GPU memory during 
training, while HAT-Net with GELU \cite{hendrycks2016gaussian} occupies 23.8GB GPU memory.
Hence, HAT-Net with SiLU \cite{elfwing2018sigmoid} can achieve
slightly better performance with less GPU memory consumption.

\myPara{Settings of $G_1$ and $G_2$.}
In HAT-Net, the parameters $G_1$ and $G_2$ play pivotal roles, controlling grid sizes for local attention calculation and downsampling rates for global attention calculation, respectively. In this evaluation, we assess the model's performance under various configurations of $G_1$ and $G_2$. By default, for tasks such as object detection and instance segmentation, we employ $G_1={8,8,8}$ and $G_2={16,8,4}$ for the $t$-th stage ($t={2,3,4}$), respectively. Subsequently, we systematically vary $G_1$ and $G_2$, evaluating the performance of Mask R-CNN \cite{he2017mask} with HAT-Net-Small as the backbone. The evaluation results, conducted on the MS-COCO \texttt{val2017} dataset \cite{lin2014microsoft}, are presented in \tabref{tab:ablation_G1G2}. The findings indicate that HAT-Net demonstrates robustness across different $G_1$ and $G_2$ settings. Notably, altering $G_1$ from its default ${8,8,8}$ configuration has a marginal impact on performance, resulting in slight performance reduction. Similarly, adjusting the values of $G_2$ yields a trade-off: decreasing values enhances performance at the expense of increased computational cost, while increasing values reduces computational cost at the cost of slightly degraded performance. Our default choice of $G_1={8,8,8}$ and $G_2={16,8,4}$ strikes a favorable balance between accuracy and efficiency, offering a practical configuration for general use.

\section{Conclusion}
This paper addresses the inefficiency inherent in vanilla vision transformers due to the elevated computational and space complexity associated with MHSA. In response to this challenge, we introduce a novel hierarchical framework for MHSA computation, denoted as H-MHSA, aiming to alleviate the computational and space demands. Compared to existing approaches in this domain, such as PVT \cite{wang2021pyramid} and Swin Transformer \cite{liu2021swin}, H-MHSA distinguishes itself by directly capturing both global dependencies and local relationships. Integrating the proposed H-MHSA, we formulate the HAT-Net family, showcasing its prowess through comprehensive experiments spanning image classification, semantic segmentation, object detection, and instance segmentation. Our results affirm the efficacy and untapped potential of HAT-Net in advancing representation learning.

\myPara{Applications of HAT-Net.}
The versatility of HAT-Net extends its utility across diverse real-world scenarios and downstream vision tasks. As a robust backbone network for feature extraction, HAT-Net seamlessly integrates with existing prediction heads and decoder networks, enabling proficient execution of various scene understanding tasks. Furthermore, HAT-Net's adaptability to different input resolutions and computational resource constraints is facilitated by the flexible adjustment of parameters, specifically $G_1$ and $G_2$. Users can tailor HAT-Net to their specific requirements, selecting from different HAT-Net versions to align with their objectives.

In conclusion, HAT-Net not only presents a pragmatic solution to the limitations of vanilla vision transformers but also opens avenues for innovation in the future design of such architectures. The simplicity of the proposed H-MHSA underscores its potential as a transformative element in the evolving landscape of vision transformer development.

\backmatter

\bibliographystyle{IEEEtran}
\bibliography{reference}

\end{document}